\documentclass[
]{ceurart}

\usepackage{listings}
\usepackage{multirow}
\usepackage{graphicx}
\begin{document}

\copyrightyear{2026}
\copyrightclause{Copyright for this paper by its authors.
  Use permitted under Creative Commons License Attribution 4.0
  International (CC BY 4.0).}

\conference{CLEF 2026 Working Notes,
  September 21 -- 24, 2026,  Jena, Germany}

\title{Through the Eyes of the Beholder: Biometric and Demographic Conditioning for Multimodal Sexism Detection}

\title[mode=sub]{Notebook for the EXIST Lab at CLEF 2026}

% \tnotetext[1]{You can use this document as the template for preparing your
%   publication. We recommend using the latest version of the ceurart style.}

\author[1]{Ana-Maria Luisa Mocanu}[%
orcid=0009-0005-2646-4086,
email=ana_maria.mogoase@upb.ro]
% \cormark[1]
\fnmark[1]
\address[1]{National University of Science and Technology POLITEHNICA Bucharest, Splaiul Independen\c{t}ei 313, Bucure\c{s}ti 060042,
Romania}
\address[2]{Academy of Romanian Scientists, Ilfov 3, Bucharest, 050044, Romania}

\author[1]{Sebastian Mocanu}[%
orcid=0009-0007-0313-4724,
email=sebastian.mocanu@upb.ro
]
\fnmark[1]

\author[1,2]{Ciprian-Octavian Truic{\u{a}}}[%
orcid=0000-0001-7292-4462,
email=ciprian.truica@upb.ro,
url=https://sites.google.com/view/ciprian-octavian-truica,
]

\author[1]{Elena-Simona Apostol}[%
orcid=0000-0001-6397-4951,
email=elena.apostol@upb.ro,
url=https://sites.google.com/view/elena-simona-apostol,
]
\cormark[1]

\cortext[1]{Corresponding author.}
\fntext[1]{These authors contributed equally.}

\begin{abstract}
Detecting sexism on the internet is a fundamentally subjective task; our team, \textbf{VANGUARD}, addresses this challenge in the EXIST 2026 Task 2 by proposing a human-centered multimodal framework that analyses and incorporates the psychological and demographic characteristics of human annotators into the detection pipeline. We fuse five input modalities through a cross-attention architecture with Feature-wise Linear Modulation conditioning. Meme text is extracted and visually described with Gemma 4, then augmented by automatic translation between English and Spanish with NLLB-200. Text and image representations are produced by LoRA-adapted XLM-RoBERTa and CLIP encoders and fused with sensor features encoded by a pretrained autoencoder. To model annotator subjectivity, we frame Subtask 2.1 as a label distribution learning problem, optimizing a Kullback-Leibler divergence loss over the full annotator label distribution. At inference time, predictions are produced by soft-voting between the deep multimodal network and a complementary SVM trained on stylometric and physiological features. Our best submission ranks 29th out of 114 on Subtask 2.2 (source intention) under soft evaluation, and the normalized ICM scores remain above the baseline on Subtasks 2.1 and 2.2, indicating that annotator-centered conditioning contributes a usable signal. We release our full pipeline and analysis to support reproducible human-centered modeling. 
\end{abstract}


\maketitle

% Participants will be required to submit their runs and will have the possibility to provide a technical report that should include a brief description of their approach, focusing on the adopted algorithms, models and resources, a summary of their experiments, and an analysis of the obtained results. Although we recommend to participate in all subtasks and in both languages, participants are allowed to participate just in one of them (e.g. subtask 2.1) and in one language (e.g. English).

\section{Introduction}

Meme culture has evolved from the early days of the internet into one of its dominant forms of communication. While often harmless, the internet's anonymity allows for the spread of controversial content \cite{isaac2018reinforcement,dutta2021reinforcement}; the subject of this paper is that the memes sometimes hide sexism \cite{arguello2023m}, often as a response against feminism. Although mostly directed at women, sexism can affect men as well \cite{duggan2017online}.

The main challenge with this type of content is its subtlety. Sexism in memes rarely appears as direct hate speech. Instead, it relies on irony (e.g., text formats like "SpongeBob mocking") and hidden visual meaning \cite{kiela2020hateful} that are hard to detect without a deep understanding of internet culture. Understanding memes remains highly subjective, as an image that one person finds offensive might be misunderstood or ignored by another, due to different personal biases and backgrounds.

EXIST 2026 task 2 addresses this challenge. Instead of forcing a single label as the ground truth, the goal is to learn from human disagreement \cite{uma2021learning,wu2023don}. By analyzing the multimodal dataset \cite{arcos2026human}, we aim to "see" through the hidden online sexism by introducing a human-centered model. Rather than looking at the meme itself, our approach incorporates real-time reaction data from annotators who saw the memes as an important component.

Our specific contributions\footnote{Code available at \url{https://github.com/DS4AI-UPB/VANGUARD-CLEF2026-EXIST}.} are as follows:
\begin{itemize}
    \item \textbf{Biometric and Demographic Conditioning:} We incorporate annotator eye-tracking, and heart rate variability features alongside demographic embeddings (gender, age, education level, ethnicity) directly into the neural model via FiLM conditioning \cite{perez2018film}, allowing the network to adjust its predictions based on the measurable physiological and social context of perception. Electroencephalography (EEG) bandpower features, while not individually significant for the neural branch, are additionally exploited by the classical SVM component of our ensemble. We report multi-seed ablations characterizing the empirical contribution of this conditioning in Section~\ref{sec:experiments}.
    
    \item \textbf{Cross-lingual Data Augmentation:} To improve cross-lingual robustness and expand the effective training set, we translate cleaned meme text and visual descriptions between English and Spanish, effectively doubling the training data while encouraging language-invariant learning.
    
    \item \textbf{VLM-based Text and Visual Enrichment:} We use the Gemma 4~\cite{gemma4_2026} vision-language model to extract clean, uncensored embedded text from memes and to generate structured natural-language descriptions of their visual components, providing richer input representations than raw pixels alone.
    
    \item \textbf{Label Distribution Learning:} For Subtask 2.1, we frame prediction as a distribution learning problem, training the model to reproduce the full distribution of annotator opinions using a soft-label Kullback-Leibler (KL) divergence \cite{Kullback1951OnIA,wu2023don}, rather than optimizing for a majority-vote binary label.
    
    \item \textbf{Cross-Attention Multimodal Fusion:} Text representations from LoRA-adapted XLM-RoBERTa~\cite{conneau2020unsupervised} and visual representations from LoRA-adapted CLIP \cite{radford2021learning,hu2022lora} are fused via cross-attention, with sensor embeddings injected through FiLM modulation \cite{perez2018film}.
    
    \item \textbf{Neural-Classical Ensemble:} Final predictions combine the deep multimodal model with an SVM trained on interpretable stylometric and physiological features via soft-voting, providing complementary coverage and acting as an algorithmic regularizer.

    \item \textbf{Ablation under Multiple-Comparison Control:} We retrain every architectural variant over five random seeds and test each against the baseline with Holm-Bonferroni correction~\cite{Holm1979ASS} across the full family of $63$ comparisons. No single component, including the biometric conditioning, survives the correction. We report this as a cautionary result.
\end{itemize}

The remainder of the paper is structured as follows. Section \ref{sec:Rw} reviews related work on sexism detection and multimodal meme analysis. Section \ref{sec:analysis} presents a statistical analysis of the EXIST 2026 dataset. Section \ref{sec:methodologies} describes our full system architecture and training strategy. Section \ref{sec:experiments} reports experimental results and ablation studies. Section \ref{sec:conclusions} concludes with a summary and the future directions.

\section{Related Work} \label{sec:Rw}

Sexism detection has evolved from text-based approaches using TF-IDF features with classical classifiers~\cite{jha2017does,de2017offensive} to deep learning with pre-trained embeddings \cite{gasparini2018multimodal}, and more recently to transformer-based models that enable deeper contextual understanding \cite{parikh2021categorizing}. The EXIST task series \cite{plaza2026overview,plaza2026overviewext,plaza2024overview,plaza2025overview} has been proven to advance the field, progressively expanding from text-only tweets to multimodal memes and videos. The 2026 edition introduces a fundamental shift from previous years. Beyond the meme content and annotator labels, the dataset includes physiological signals and demographic profiles of the annotators \cite{arcos2026human}, making it possible to use human-centered modeling approaches.

Memes pose a unique multimodal challenge, as their meaning arises from the interplay between text and image, often masked by humor or irony. The Hateful Memes Challenge \cite{kiela2020hateful} showed that state-of-the-art multimodal models struggled when offensiveness depended on the combination of individually benign modalities. Recent approaches have addressed the modality gap through VLM-based captioning \cite{cao2023pro}, multimodal fusion and bias analysis \cite{rizzi2023recognizing}, and graph-based reasoning \cite{italiani2026memeweaver}. We additionally condition the fusion on annotator biometrics through FiLM modulation~\cite{perez2018film}, and use Gemma~4~\cite{gemma4_2026} to both extract embedded meme text and generate structured visual descriptions, further augmenting the training data through cross-lingual translation with NLLB-200~\cite{costa2022no}.

Annotator disagreement in subjective tasks carries a genuine signal rather than noise~\cite{uma2021learning}. Prior work has addressed this by training models to predict soft label distributions directly~\cite{wu2023don}. We adopt this framing, weighting each instance by its annotator entropy, and additionally employ Supervised Contrastive Learning~\cite{khosla2020supervised} as an auxiliary loss to geometrically structure the embedding space alongside the primary KL divergence objective.

Eye-tracking and EEG signals have been previously used in affective computing for emotion recognition~\cite{lim2020emotion,fu2023novel}, but their integration into sexism detection introduced in EXIST 2026~\cite{arcos2026human} is unique. We encode the retained physiological features through a pretrained sensor autoencoder before joint training and inject them via FiLM modulation~\cite{perez2018film} into a cross-attention architecture.

\section{Data Analysis} \label{sec:analysis}
The EXIST 2026 dataset \cite{arcos2026human} consists of 3984 memes which were viewed by 16 subjects. Additionally, it contains human data composed of recorded eye movements, heart rate, and EEG of the annotators.

\subsection{Demographic and Physiological Analysis}
To model the multimodal indicators of sexism, we fused textual, visual, demographic, and physiological data. The ground truth was established via majority voting, where a meme was classified as sexist (i.e., $y=1$) if $\ge 50\%$ of annotators agreed. A soft-label target ($p_{yes}$) was also retained to model annotator disagreement through the soft KL Divergence objective during training. We additionally implemented an uncertainty-weighted variant of this objective, applying a per-instance factor $\exp\!\big(-p_{yes}(1-p_{yes})\big)$ that emphasizes high-consensus samples. This variant was explored in preliminary single-task experiments and is not used in the multitask runs reported in this paper.

\textbf{Demographic and Physiological Selection.} Pearson's $\chi^2$ tests confirmed that demographic traits (i.e., gender, age, study level, and ethnicity) significantly influence labeling behavior ($p < 0.001$), leading to their inclusion via trainable embedding layers. For physiological telemetry, we aggregated Eye Tracking (ET), Heart Rate (HR), and EEG data \cite{lim2020emotion,fu2023novel} by computing the mean across users per instance. Following $t$-tests and point biserial correlation analysis represented in Table \ref{tab:stat_analysis}, we retained only the features showing significant variance ($p < 0.05$).

\begin{table}[htbp]
\caption{Statistical Significance of Demographic and Physiological Features}
\label{tab:stat_analysis}
\centering
\begin{tabular}{llrc}
\hline
\textbf{Feature Category} & \textbf{Variable} & \textbf{Test Statistic} & \textbf{p-value} \\
\hline
Demographics ($\chi^2$) & Gender & $\chi^2 = 79.51$ & $< 0.001^{**}$ \\
 & Age & $\chi^2 = 187.77$ & $< 0.001^{**}$ \\
 & Study Level & $\chi^2 = 45.77$ & $< 0.001^{**}$ \\
 & Ethnicity & $\chi^2 = 41.13$ & $< 0.001^{**}$ \\
\hline
Physiological (t-test) & Reaction Time & $t = 8.17$ & $< 0.001^{**}$ \\
 & Fixations Count & $t = 5.52$ & $< 0.001^{**}$ \\
 & Saccades Count & $t = 5.44$ & $< 0.001^{**}$ \\
 & HR Std. & $t = 3.45$ & $< 0.001^{**}$ \\
 & Mean Pupil Diameter & $t = 0.00$ & $0.9962$ \\
 & Mean Heart Rate & $t = -1.16$ & $0.2466$ \\
 & EEG Alpha Power & $t = -1.01$ & $0.3123$ \\
\hline
\multicolumn{4}{l}{** Indicates statistical significance at $\alpha = 0.05$ level.}
\end{tabular}
\end{table}

\begin{figure}[htbp]
    \centering
    \includegraphics[width=1\linewidth]{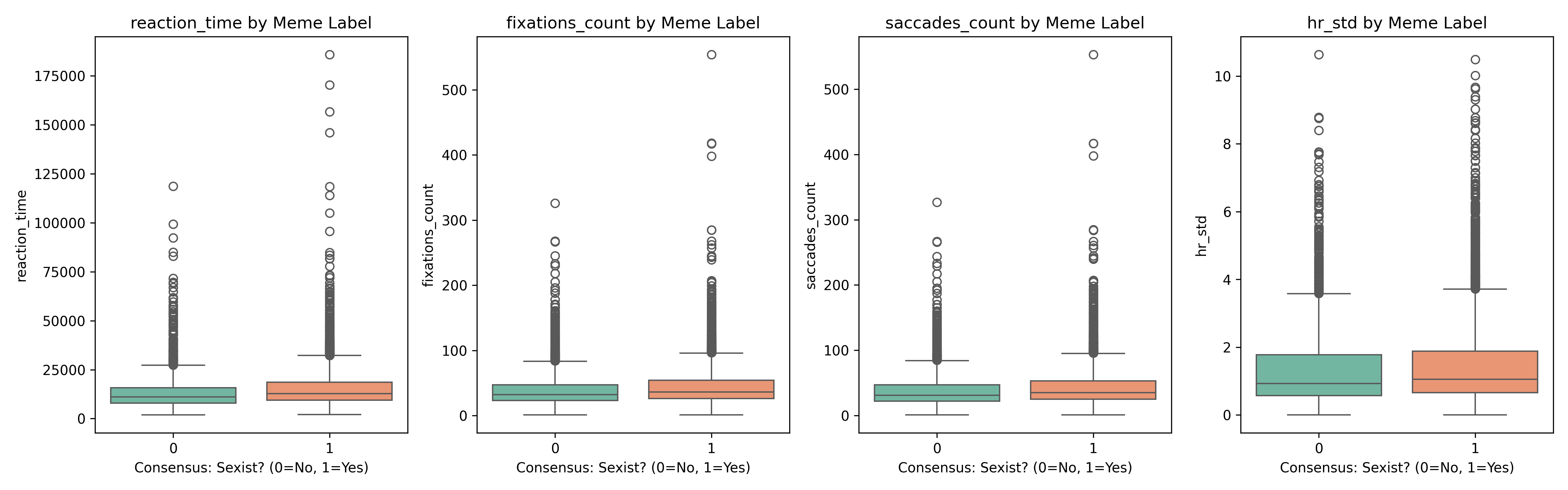}
    \caption{Distribution of predictive physiological features across meme classes.}
    \label{fig:relevant-features}
\end{figure}

\textbf{Feature Analysis.} The distribution of predictive features is visualized in Figure \ref{fig:relevant-features}. The analysis of the boxplots indicates that sexist content triggers significantly higher reaction times, fixation counts, and increased heart rate variability (HR Std.). This suggests that sexist memes generate greater cognitive friction than non-sexist ones.

\begin{figure}[htbp]
    \centering
    \includegraphics[width=0.8\linewidth]{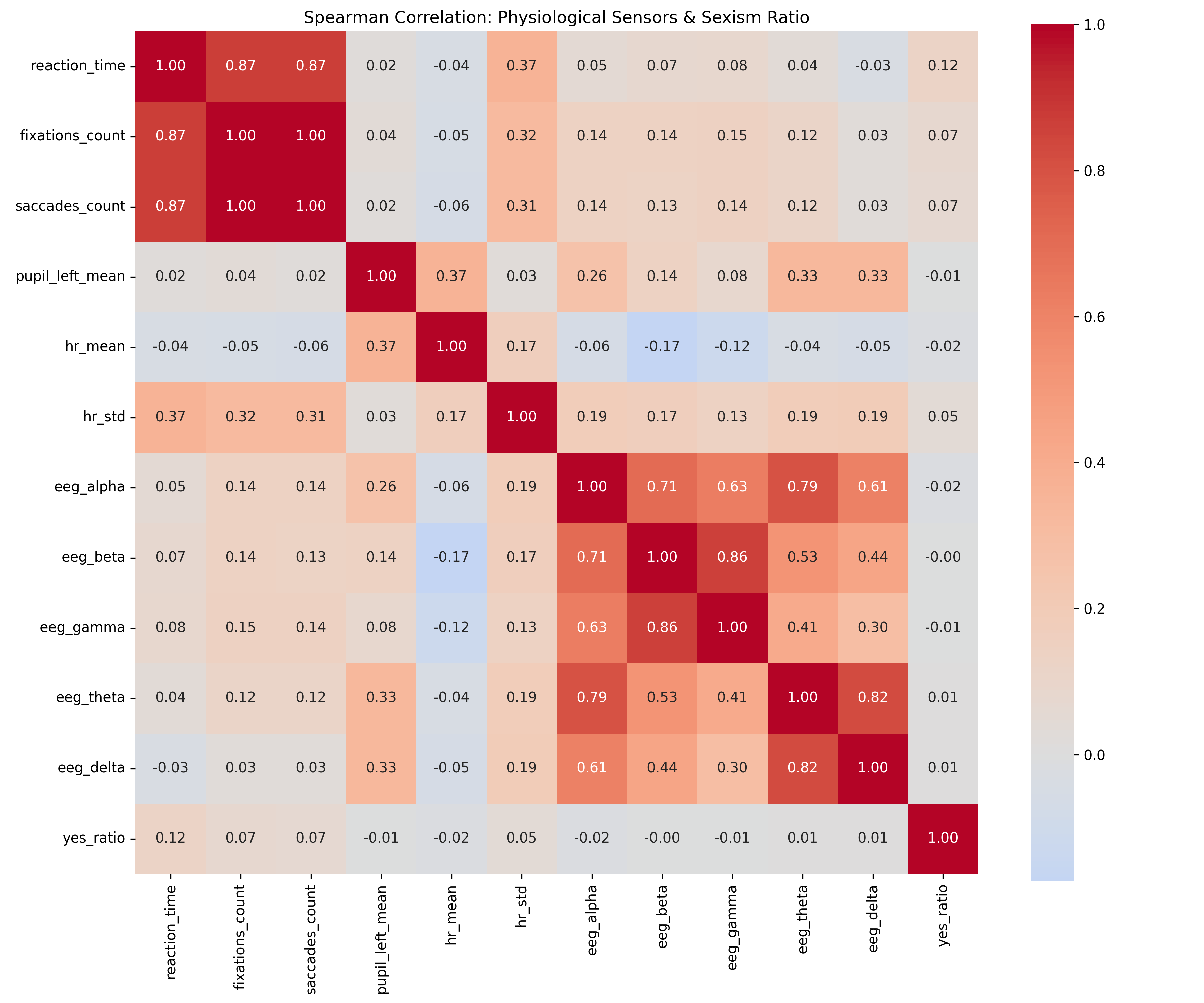}
    \caption{Spearman correlation heatmap of aggregated physiological sensors.}
    \label{fig:heatmap-sensors}
\end{figure}

Despite their predictive power, the Spearman correlation heatmap represented in Figure \ref{fig:heatmap-sensors} reveals extreme multicollinearity among eye-tracking metrics, with Fixations and Saccades yielding $\rho \approx 1.0$. While this indicates functional redundancy, all significant features were retained to allow the neural network to optimize weighting internally.

\textbf{Network Integration.} To prevent features with large numerical scales, such as reaction time, from dominating the gradients, the 4-dimensional sensor vector was standardized using a \texttt{StandardScaler} fitted on the training split and applied to the validation and test data. Reaction time was additionally log-transformed prior to scaling to compress its heavy right tail. This ensures that subtle fluctuations in heart rate variance are weighed equitably alongside high-magnitude eye-tracking durations during multimodal fusion.

\subsection{Text}
To capture the linguistic nuances of the sexist memes, we extracted metrics from the embedded text across $N=3,984$ valid records before preprocessing.

\textbf{Linguistic Distribution.} The dataset presents a highly balanced bilingual composition, with the records split almost perfectly between English ($n=2,005$) and Spanish ($n=1,979$). This is crucial for training, ensuring the network does not falsely correlate a specific language with the presence of sexism.

\textbf{Text Length Analysis.} Memes inherently rely on brief, punchy text and prominent visual cues, occasionally exhibiting longer blocks of text. As detailed in Table \ref{tab:text_stats}, the analysis of the extracted textual fields reveals an average character count of $124.3$ ($\sigma=104.72$) and a mean word count of $21.6$ words ($\sigma=18.12$). The word distributions are right-skewed. The maximum word count reaches 299 words, while $75\%$ of the dataset contains 26 words or fewer.

\begin{table}[htbp]
\caption{Descriptive Statistics of Textual Features ($N=3,984$)}
\label{tab:text_stats}
\centering
\begin{tabular}{lrrrrrr}
\hline
\textbf{Feature} & \textbf{Mean} & \textbf{Std. Dev.} & \textbf{Min} & \textbf{25\%} & \textbf{Median} & \textbf{Max} \\
\hline
Character Count & 124.30 & 104.72 & 9.00 & 64.00 & 96.00 & 1777.00 \\
Word Count & 21.60 & 18.12 & 2.00 & 11.00 & 17.00 & 299.00 \\
\hline
\end{tabular}
\end{table}

\subsection{Image}
For the visual modality, exploratory data analysis was conducted on the structural and pixel-level properties of the raw image.

\begin{figure}[htbp]
    \centering
    \includegraphics[width=1\linewidth]{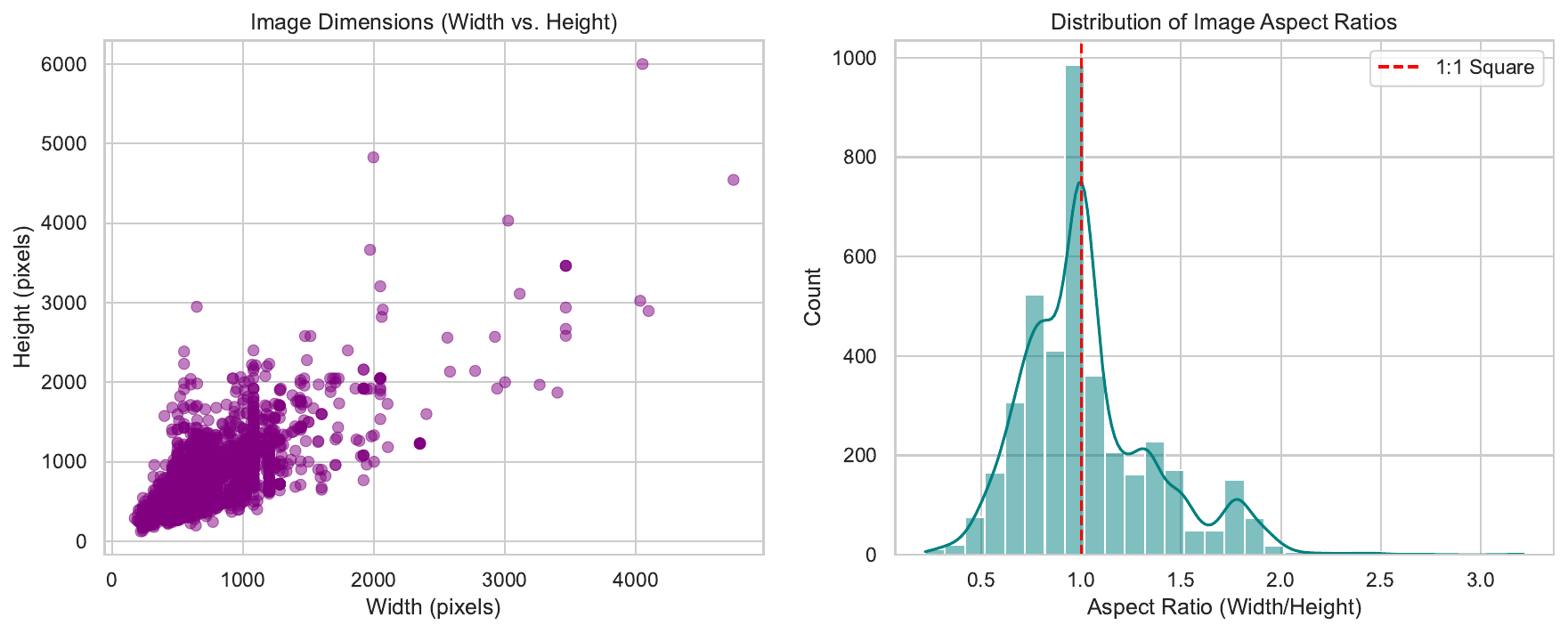}
    \caption{Image dimensions (width vs. height) and the resulting aspect ratio distribution, highlighting a dominant $1:1$ structural format.}
    \label{fig:image-features}
\end{figure}

\textbf{Dimension and Formatting.} The resolution of the memes varies significantly, with widths ranging from 173 to 4744 pixels and heights from 124 to 6000 pixels. The aspect ratio distribution shown in Table~\ref{tab:image_stats} exhibits a distinct median exactly at 1 and a mean of 1.03. As represented in Figure \ref{fig:image-features}, the dominance of the 1:1 square format is characteristic of meme formats. Standardizing the input via center-cropping or padding to a fixed square resolution results in minimal information loss. However, this affects taller memes, typically those with a storytelling format, which may benefit from being split into multiple sub-images before being fed as input.

\textbf{Visual Tones. } An analysis of the average grayscale pixel brightness, ranging from $0 = \text{Black}$ to $255 = \text{White}$, revealed an average brightness of $\mu=138.31$ ($\sigma=43.55$), centered perfectly in mid-tones, with the interquartile range falling between $108.35$ and $168.68$.

\begin{table}[htbp]
\caption{Descriptive Statistics of Visual Features ($N=3984$)}
\label{tab:image_stats}
\centering
\begin{tabular}{lrrrrrr}
\hline
\textbf{Feature} & \textbf{Mean} & \textbf{Std. Dev.} & \textbf{Min} & \textbf{25\%} & \textbf{Median} & \textbf{Max} \\
\hline
Image Width (px) & 713.41 & 355.18 & 173.00 & 500.00 & 640.00 & 4744.00 \\
Image Height (px) & 745.33 & 405.74 & 124.00 & 480.00 & 650.00 & 6000.00 \\
Aspect Ratio (W/H) & 1.03 & 0.34 & 0.22 & 0.80 & 1.00 & 3.22 \\
Pixel Brightness & 138.31 & 43.55 & 16.19 & 108.35 & 136.03 & 251.79 \\
\hline
\end{tabular}
\end{table}

\section{Methodologies} \label{sec:methodologies}
\begin{figure}[htbp]
\centering
\includegraphics[width=1\linewidth]{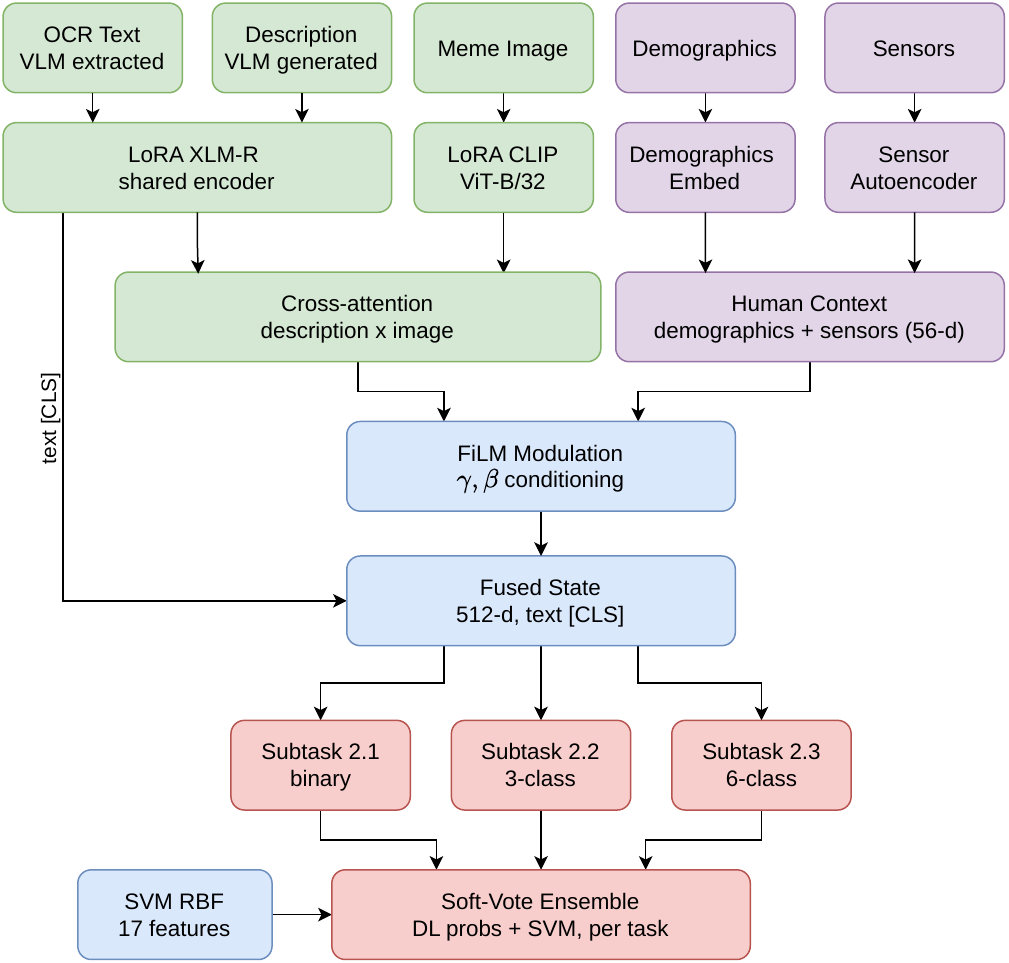}
\caption{Overview of our architecture. Five input streams, the VLM-extracted embedded text, the VLM-generated visual description, the meme RGB image, the annotator demographics, and the physiological sensor vector, are encoded by a shared LoRA XLM-RoBERTa for embedded text and description, a LoRA CLIP vision encoder for images, demographic embedding layers, and a pretrained sensor autoencoder. The projected description and image-patch sequences are fused by multi-head cross-attention. The resulting representation is modulated through FiLM by the $56$-dimensional human-context vector formed from demographics concatenated with the sensor embedding, and is then concatenated with the text \texttt{[CLS]} token to form the $512$-dimensional fused state shared across subtasks. Three task heads (binary for Subtask 2.1, three-class for 2.2, six-class multi-label for 2.3) produce per-task predictions, which at inference are combined with a feature-based SVM through soft-voting. Color denotes the meme-content path (green), the annotator-context path (purple), and the task outputs (red).}
\label{fig:architecture}
\end{figure}

Our system is a multi-stage pipeline. We first enrich each meme with clean embedded text and a structured visual description extracted by a vision-language model (see Subsection~\ref{ssec:text-enhancer}), then expand the training set through cross-lingual translation (see Subsection~\ref{ssec:data-augmentation}). The enriched data feeds a deep multimodal network which fuses five input streams through cross-attention and FiLM conditioning (see Subsection~\ref{ssec:fusion}), whose physiological branch is warm-started by a pretrained autoencoder (Subsection~\ref{ssec:auto-encoder}) and whose embedding space is shaped by an auxiliary contrastive objective (Subsection~\ref{ssec:contrastive-learning}). At inference time, the network is combined with a feature-based SVM through soft-voting (Subsection~\ref{ssec:ensembling}). An overview of the complete pipeline is shown in Figure~\ref{fig:architecture}.

\subsection{Text Enhancer}
\label{ssec:text-enhancer}
The textual data and the visual descriptions generation were extracted using the Gemma 4~\cite{gemma4_2026} vision-language model via the Ollama framework. To ensure efficient processing, we used the 4-bit quantized version Effective-4B (e4b) of the model. The inference settings were kept strict, with a $0.0$ temperature and a $0.1$ top\_p, to be able to extract the literal meaning of the text and avoid hallucinations. A fallback re-extraction pass for the items whose JSON output failed to parse used the same strict settings for text transcription but slightly relaxed sampling with a temperature of $0.4$ for the visual descriptions.

\subsection{Data Augmentation}
\label{ssec:data-augmentation}
To address the limitation of the training set size and to improve the cross-lingual robustness, we performed translation-based data augmentation using the NLLB-200 distilled model. For each meme, the cleaned text and the visual description were translated into the other language (e.g., English to Spanish), and the translated copies were appended to the training set with an "\_aug" suffix on their identifiers. This approach doubles the available training data and helps the encoders learn language-invariant representations of sexist content. To prevent leakage, training and validation splits were assigned by parent ID on the original de-duplicated dataset, with each augmented copy placed in the same split as its source meme~\cite{costa2022no}.

\subsection{Sensor Auto Encoder}
\label{ssec:auto-encoder}
The four retained physiological features (i.e., log reaction time, fixation count, saccade count, and heart-rate standard deviation) are low-dimensional and noisy, which makes them prone to being overwhelmed by the high-capacity text and vision encoders during joint training. To obtain a more robust and informative representation, we pretrain a sensor autoencoder in an unsupervised manner prior to the main multimodal training stage.

The autoencoder is a symmetric multilayer perceptron which maps the 4-dimensional standardized sensor vector to a 32-dimensional latent representation through a $4 \rightarrow 64 \rightarrow 32$ stack with Layer Normalization, GELU activations~\cite{hendrycks2016gaussian}, and dropout. The decoder mirrors this as $32 \rightarrow 64 \rightarrow 4$ to reconstruct the input. The network is trained for 50 epochs with the Adam optimizer~\cite{kingma2014adam} and a mean-squared-error reconstruction objective. Once pretrained, the encoder weights are transferred into the corresponding sensor branch of the multimodal model, where they continue to be finetuned end-to-end. This warm start provides the fusion network with a meaningful 32-dimensional physiological embedding from the first training step, rather than forcing it to learn one from scratch alongside the much larger language and vision components.

\subsection{Cross-Attention Fusion} 
\label{ssec:fusion}
Our model fuses five input streams: the meme embedded text (extracted via VLM), the VLM-generated visual description, the meme image, annotator demographics, and the physiological sensor vector.

\textbf{Encoders.} The embedded text and the visual description are independently encoded by a shared LoRA-adapted XLM-RoBERTa \cite{conneau2020unsupervised}, while the image is encoded by a LoRA-adapted CLIP ViT-B/32 vision model \cite{radford2021learning,hu2022lora}. LoRA adapters (rank 16, $\alpha=32$) are inserted into the attention projections of both encoders, keeping the pretrained backbones frozen and training only a small number of additional parameters. The token sequences from each encoder are linearly projected to a common 256-dimensional space with Layer Normalization~\cite{ba2016layer} and dropout.

\textbf{Grounding the description in the image.} To resolve the meaning of a meme, which often emerges from the interplay between what is written and what is shown, we ground the textual description in the visual content through a multi-head cross-attention layer~\cite{vaswani2017attention}. The projected description sequence serves as the query, and the projected image patch sequence serves as the keys and values. A residual connection followed by Layer Normalization~\cite{ba2016layer} yields a fused vision-language sequence, from which we take the \texttt{[CLS]} position as the fused representation $h_{\text{fused}} \in \mathbb{R}^{256}$. 

\textbf{Human conditioning via FiLM.} Annotator demographics are passed through dedicated embedding layers (i.e., gender, age, study level, ethnicity) and averaged across valid annotations of each meme to form a demographic vector. This is concatenated with the 32-dimensional physiological embedding produced by the pretrained sensor encoder (Subsection~\ref{ssec:auto-encoder}), giving a 56-dimensional human-context vector. Following FiLM \cite{perez2018film}, two linear layers map this vector to per-feature scale $\gamma$ and shift $\beta$ parameters, which modulate the fused representation as $h_{\text{mod}} = h_{\text{fused}} \odot (1 + \gamma) + \beta$. Both FiLM projections are zero-initialized, so the model begins training as an unconditioned multimodal classifier and progressively learns how much to let annotator context reshape its predictions.

\textbf{Classification Heads.} The modulated representation is concatenated with the text \texttt{[CLS]} embedding to form the final 512-dimensional fused state, which is shared across all subtasks. Each active subtask has its own classification head: a binary head for Subtask 2.1, a three-class head for Subtask 2.2, and a six-class multi-label head for Subtask 2.3. For Subtask 2.1, an auxiliary binary sexism head is attached to $h_{\text{fused}}$ to provide an additional training signal, and a 128-dimensional projection of the final fused state feeds the supervised contrastive objective.

\subsection{Contrastive Learning}
\label{ssec:contrastive-learning}
In addition to the primary KL divergence objective, we employ Supervised Contrastive Learning (SupCon) \cite{khosla2020supervised} as an auxiliary loss to shape the embedding space. This operates on a 128-dimensional projection of the final fused state as in Equation \eqref{eq:contrastive}, where $z_i$ are L2-normalized embeddings, $P(i)$ denotes the set of positive pairs, and $\tau = 0.07$ is the temperature parameter.

This addition complements the KL divergence objective by explicitly structuring the embedding space. Memes with the same consensus label are pulled together while those with differing labels are pushed apart. The dual optimization encourages representations that are simultaneously distributionally calibrated and geometrically well-separated.

\begin{equation}
    \mathcal{L}_{\text{SupCon}} = -\frac{1}{|P(i)|} \sum_{p \in P(i)} \log \frac{\exp(\text{sim}(z_i, z_p) / \tau)}{\sum_{a \neq i} \exp(\text{sim}(z_i, z_a) / \tau)}
\label{eq:contrastive}
\end{equation}

\subsection{Ensembling}
\label{ssec:ensembling}
The final prediction system combines two complementary classifiers through soft-voting. The first is our deep multimodal model, a LoRA-adapted neural network described in Subsection \ref{ssec:fusion} that processes all five input modalities through cross-attention and FiLM conditioning \cite{perez2018film}. The second is a feature-based Support Vector Machine (SVM) with an RBF kernel, trained on a 17-dimensional feature vector comprising text stylometry (i.e., length, word count, capitalization ratio, punctuation ratio), demographic ratios (e.g., female and older-annotator proportions), and log-scaled physiological signals (i.e., eye-tracking, heart rate, and EEG bandpower). For Subtask 2.1 the SVM is a binary classifier, for Subtask 2.2 a multiclass classifier, and for Subtask 2.3 a One-vs-Rest classifier producing one binary decision per category. All classifiers use balanced class weights. The ensemble prediction is computed in Equation~\eqref{eq:ensemble}, where the blending weight $\alpha$ and the decision threshold $\tau$ are jointly optimized via grid search on the validation set to maximize the macro F1-score. The SVM acts as an algorithmic regularizer, operating on interpretable handcrafted features and providing a complementary signal that can correct neural model errors.

\begin{equation}
    \hat{p} = \alpha \cdot p_{\text{DL}} + (1 - \alpha) \cdot p_{\text{SVM}}
\label{eq:ensemble}
\end{equation}

\section{Experiments} \label{sec:experiments}

\subsection{Experimental Setup}
All the experiments were conducted on a single NVIDIA GPU with PyTorch 2.x. 
For the text augmentation and meme visual interpretation, Gemma 4~\cite{gemma4_2026} was used. 
The text encoder is XLM-RoBERTa-base, and the image encoder is CLIP ViT-B/32, both adapted with LoRA, using rank $16$, $\alpha$ $32$, and a $0.1$ dropout.
Training used AdamW with differential learning rates~\cite{loshchilov2017decoupled}, $3 \times 10^{-5}$ for fusion heads and $8 \times 10^{-6}$ for LoRA parameters, with weight decay of $0.1$. Gradients were accumulated over $4$ steps and clipped to norm $1$.

\subsection{Results}

Tables~\ref{tab:soft-all} and~\ref{tab:hard-all} report our official EXIST 2026 results under the soft and hard evaluation protocols, respectively, broken down by subtask and language split, alongside the corresponding validation scores computed on the held-out 20\% of the training set with PyEvALL. For reference, each table also lists the organizers majority and minority class baselines on the All split.

Our system is most competitive on Subtask 2.2 (i.e., source intention), where it ranks 29th of 114 systems under soft evaluation, indicating that the fused multimodal representation captures the distinction between direct and judgmental sexism reasonably well. Performance on Subtask 2.1 (i.e., binary sexism detection) is mid-range, with the English split ranked 86th of 214 with hard ICM-Norm $0.5496$ and F1 $0.7336$ outperforming the Spanish split, which is ranked 141st of 214 with hard ICM-norm $0.4306$ and F1 $0.6697$ on test. This English-Spanish gap is much less pronounced on validation, where the two splits perform similarly under hard evaluation (ICM-Norm $0.5609$ vs. $0.4679$), and Spanish even exceeds English on the YES-class F1 ($0.7732$ vs. $0.7556$). The widening of the gap from validation to test suggests that the Spanish portion of the test distribution diverges more from the training distribution than the English portion does, plausibly because the multilingual pretraining of XLM-RoBERTa and CLIP captures English internet sexism cues more idiomatically than Spanish ones. Subtask 2.3 (i.e., fine-grained categorization) is the hardest setting, reflecting the difficulty of the six-way multi-label problem under severe class imbalance, where rare categories receive few positive annotations.

Relative to the organizers baselines, our submissions clear the majority-class system by a wide margin on Subtasks 2.1 and 2.2 under both protocols (e.g., $0.4861$ vs. $0.2947$ hard ICM-Norm on Subtask 2.1, and $0.3612$ vs $0.1369$ on Subtask 2.2), confirming that the human-centered conditioning contributes a usable signal rather than noise. Subtask 2.3 is the exception, although the soft score is $0.1568$ still exceeds the majority class baseline which is $0$, the hard scores collapses onto that baseline exactly with $0.0703$ ICM-Norm.

\begin{table*}[h]
\centering
\caption{Soft evaluation results for Tasks 2.1, 2.2, and 2.3. For each language split the best submitted run is reported. The test and validation columns are each selected for their own best score and may therefore originate from different runs.}
\label{tab:soft-all}
\resizebox{\textwidth}{!}{%
\begin{tabular}{llccccccc}
\hline
\textbf{Task} & \textbf{Split} & \textbf{Rank} & \textbf{ICM-Soft} & \textbf{ICM-Soft Norm} & \textbf{CE} & \textbf{Val. ICM-Soft} & \textbf{Val. ICM-Soft Norm} & \textbf{Val. CE} \\
\hline
\multirow{3}{*}{2.1} & All & 75  & $-0.5545$ & $0.4109$ & $0.9389$ & $-0.5584$ & $0.4126$ & $0.9414$ \\
                     & EN  & 75  & $-0.4360$ & $0.4292$ & $0.9200$ & $-0.4013$ & $0.4372$ & $0.9255$ \\
                     & ES  & 70  & $-0.6843$ & $0.3909$ & $0.9568$ & $-0.7408$ & $0.3842$ & $0.9570$ \\
\hline
\multirow{3}{*}{2.2} & All & 29  & $-1.5152$ & $0.3389$ & $1.4575$ & $-1.4622$ & $0.3475$ & $1.4564$       \\
                     & EN  & 30  & $-1.3087$ & $0.3572$ & $1.4479$ & $-1.1936$ & $0.3734$ & $1.4360$       \\
                     & ES  & 31  & $-1.7624$ & $0.3169$ & $1.4667$ & $-1.7647$ & $0.3188$ & $1.4764$       \\
\hline
\multirow{3}{*}{2.3} & All & 41  & $-6.4755$ & $0.1568$ & --       & $-6.6050$ & $0.1550$ & --   \\
                     & EN  & 41  & $-6.2800$ & $0.1607$ & --       & $-6.4860$ & $0.1590$ & --   \\
                     & ES  & 38  & $-6.6920$ & $0.1525$ & --       & $-6.7264$ & $0.1505$ & --   \\
\hline
\multicolumn{9}{l}{\textit{Organizer baselines (test set, All split)}} \\
\hline
2.1 & Maj.-class & 140 & $-2.3568$  & $0.1212$ & $4.4015$ & -- & -- & -- \\
2.1 & Min.-class & 141 & $-3.5089$  & $0.0000$ & $5.5672$ & -- & -- & -- \\
2.2 & Maj.-class & 112 & $-5.0745$  & $0.0000$ & $5.5565$ & -- & -- & -- \\
2.2 & Min.-class & 114 & $-18.9382$ & $0.0000$ & $8.0245$ & -- & -- & -- \\
2.3 & Maj.-class & 74  & $-9.8173$  & $0.0000$ & --       & -- & -- & -- \\
2.3 & Min.-class & 114 & $-50.0353$ & $0.0000$ & --       & -- & -- & -- \\
\hline
\end{tabular}}
\end{table*}

\begin{table*}[h]
\centering
\caption{Hard evaluation results for Tasks 2.1, 2.2, and 2.3. For each language split the best submitted run is reported. The test and validation columns are each selected for their own best score and may therefore originate from different runs.}
\label{tab:hard-all}
\resizebox{\textwidth}{!}{%
\begin{tabular}{llccccccc}
\hline
\textbf{Task} & \textbf{Split} & \textbf{Rank} & \textbf{ICM-Hard} & \textbf{ICM-Hard Norm} & \textbf{F1} & \textbf{Val. ICM-Hard} & \textbf{Val. ICM-Hard Norm} & \textbf{Val. F1} \\
\hline
\multirow{3}{*}{2.1} & All & 117 & $-0.0274$           & $0.4861$ & $0.6920$ & $\phantom{-}0.0410$ & $0.5210$ & $0.7651$ \\
                     & EN  & 86  & $\phantom{-}0.0977$ & $0.5496$ & $0.7336$ & $\phantom{-}0.1208$ & $0.5609$ & $0.7556$ \\
                     & ES  & 141 & $-0.1362$           & $0.4306$ & $0.6697$ & $-0.0610$           & $0.4679$ & $0.7732$ \\
\hline
\multirow{3}{*}{2.2} & All & 71  & $-0.3992$           & $0.3612$ & $0.3719$ & $-0.3581$           & $0.3709$ & $0.4040$ \\
                     & EN  & 60  & $-0.3303$           & $0.3854$ & $0.3825$ & $-0.3588$           & $0.3714$ & $0.4017$ \\
                     & ES  & 69  & $-0.4367$           & $0.3479$ & $0.3644$ & $-0.3662$           & $0.3655$ & $0.4033$ \\
\hline
\multirow{3}{*}{2.3} & All & 138 & $-2.0711$           & $0.0703$ & $0.0919$ & $-1.8981$           & $0.1159$ & $0.1361$ \\
                     & EN  & 143 & $-2.0015$           & $0.0747$ & $0.0938$ & $-1.8276$           & $0.1164$ & $0.1389$ \\
                     & ES  & 130 & $-2.1173$           & $0.0667$ & $0.0901$ & $-1.9635$           & $0.1148$ & $0.1332$ \\
\hline
\multicolumn{9}{l}{\textit{Organizer baselines (test set, All split)}} \\
\hline
2.1 & Maj.-class & 202 & $-0.4038$ & $0.2947$ & $0.6821$ & -- & -- & -- \\
2.1 & Min.-class & 214 & $-0.6468$ & $0.1711$ & $0.0000$ & -- & -- & -- \\
2.2 & Maj.-class & 167 & $-1.0445$ & $0.1369$ & $0.1839$ & -- & -- & -- \\
2.2 & Min.-class & 183 & $-2.0637$ & $0.0000$ & $0.0697$ & -- & -- & -- \\
2.3 & Maj.-class & 140 & $-2.0711$ & $0.0703$ & $0.0919$ & -- & -- & -- \\
2.3 & Min.-class & 174 & $-3.3135$ & $0.0000$ & $0.0318$ & -- & -- & -- \\
\hline
\end{tabular}}
\end{table*}

\subsection{Augmentation and Ensemble Ablation}
\label{subsec:augmentation-ensemble-ablation}
Table~\ref{tab:ablation} compares four configurations on Subtask 2.1, isolating the contributions of cross-lingual augmentation and the SVM ensemble. Three of these correspond to our three official submissions. The augmentation-only variant without ensemble was retained for ablation purposes and was not submitted due to limited submission numbers.

\begin{table}[htbp]
\caption{Ablation study on Subtask 2.1 validation (i.e., subset of training set) and test (i.e., results from the submitted runs) sets with hard evaluation. All three configurations correspond to officially submitted runs. \textit{Aug.}\ indicates whether cross-lingual translation augmentation was used during training and \textit{Ens.}\ indicates whether the SVM ensemble was applied at inference. The augmented without ensemble configuration was not submitted for official evaluation.}
\label{tab:ablation}
\centering
\begin{tabular}{lllcccccc}
\toprule
& \multicolumn{2}{c}{\textbf{Configuration}} & \multicolumn{3}{c}{\textbf{ICM-Hard Norm}} & \multicolumn{3}{c}{\textbf{F1 Yes}} \\
\cmidrule(lr){2-3} \cmidrule(lr){4-6} \cmidrule(lr){7-9}
\textbf{Dataset} & \textbf{Aug.} & \textbf{Ens.} & All & EN & ES & All & EN & ES \\
\midrule
\multirow{4}{*}{Validation} 
                            & --         & -- & $\mathbf{0.5210}$ & $\mathbf{0.5609}$ & $0.4679$ & $\mathbf{0.7651}$ & $\mathbf{0.7556}$ & $\mathbf{0.7732}$ \\
                            & \checkmark & \checkmark & $0.4963$ & $0.5001$ & $\mathbf{0.4925}$ & $0.7395$ & $0.7409$ & $0.7382$ \\
                            & --         & \checkmark & $0.4963$ & $0.5494$ & $0.4249$ & $0.7418$ & $0.7277$ & $0.7532$ \\
                            & \checkmark & --         & $0.4561$ & $0.4540$ & $0.4582$ & $0.7463$ & $0.7449$ & $0.7477$ \\
\midrule
\multirow{3}{*}{Test}       & --         & -- & $\mathbf{0.4861}$ & $0.5414$ & $\mathbf{0.4306}$ & $0.6920$ & $0.7167$ & $\mathbf{0.6697}$ \\
                            & \checkmark & \checkmark & $0.4712$ & $\mathbf{0.5496}$ & $0.3929$ & $\mathbf{0.6924}$ & $\mathbf{0.7336}$ & $0.6545$ \\
                            & --         & \checkmark & $0.4657$ & $0.5235$ & $0.4079$ & $0.6766$ & $0.7012$ & $0.6540$ \\
                            & \checkmark & --         & -- & -- & -- & -- & -- & -- \\
\bottomrule
\end{tabular}
\end{table}

The pure deep learning model trained on the non-augmented corpus achieves the highest overall scores on validation with ICM-Norm $0.5210$ and F1 Yes $0.7651$, outperforming every variant that adds either augmentation or the SVM ensemble. The same configuration also gives the best test ICM-Norm on the All split with $0.4861$. 

Cross-lingual augmentation produces a clear language trade-off. Comparing the non-augmented ensemble against the augmented ensemble on validation, augmentation lifts Spanish performance (i.e., $0.4249 \rightarrow 0.4925$ ICM-Norm) while degrading English (i.e., $0.5494 \rightarrow 0.5001$). The same pattern repeats on test, where the augmented ensemble achieves our best English score with ICM-Norm $0.5496$ and F1 $0.7336$ while the non-augmented variants are stronger on Spanish. Augmentation therefore acts as a transfer mechanism, lifting the weaker language at the cost of the stronger one.

The SVM ensemble does not provide a uniform improvement, its effect on validation depends on the training regime. Adding the ensemble to the pure deep learning system lowers performance in the no-augmentation setup ($0.5210 \rightarrow 0.4963$ ICM-Norm) but raises it in the augmentation setup ($0.4561 \rightarrow 0.4963$ ICM-Norm). However, on test the effect partially reverses, where the ensemble paired with augmentation yields the best English score, suggesting that the SVM contributes complementary signal precisely when the deep model has been exposed to noisier translated data. The ensemble is therefore best understood as a regularizer that helps in the noisier training regime but adds variance when the underlying DL system is already well fitted. This combination of effects explains why no single submission dominates across all language splits, and why the best test scores per split in Tables~\ref{tab:soft-all} and~\ref{tab:hard-all} come from different submitted runs.

\subsection{Architectural Component Ablation} 
\label{subsec:architectural-component-ablation}
To assess the contribution of the individual neural-network components, we retrained the system from scratch with each component independently disabled, repeating the procedure with five distinct random seeds per configuration. We probe the contribution of FiLM-based human conditioning, the visual description and image streams both individually and jointly (i.e., ``text-only'' variant), the supervised contrastive auxiliary loss, the auxiliary sexism head, and the multi-task formulation itself with single-task variants trained on a single subtask. All ablation runs use the non-augmented training corpus and the deep model alone without the SVM ensemble, matching the configuration of our best-performing official submission on the All split. Table~\ref{tab:comp_ablation_hard} reports the mean and standard deviation of hard-evaluation scores across the five seeds, while the Table~\ref{tab:comp_ablation_soft} reports the soft evaluation.

\begin{table}[htbp]
\caption{Component ablation on the validation set, hard evaluation on both English and Spanish split. Each cell shows mean $\pm$ standard deviation over five random seeds. All configurations trained on the non-augmented dataset using only the deep learning model.}
\label{tab:comp_ablation_hard}
\centering
\resizebox{\columnwidth}{!}{%
\begin{tabular}{lcccccc}
\hline
\multirow{2}{*}{\textbf{Configuration}} & \multicolumn{2}{c}{\textbf{Task 2.1}} & \multicolumn{2}{c}{\textbf{Task 2.2}} & \multicolumn{2}{c}{\textbf{Task 2.3}} \\
 & ICM-Norm & F1 & ICM-Norm & F1 & ICM-Norm & F1 \\
\hline
Full system (baseline) & $0.4877 \pm 0.0066$ & $0.6607 \pm 0.0054$ & $0.3517 \pm 0.0158$ & $0.3916 \pm 0.0105$ & $0.1174 \pm 0.0018$ & $0.1339 \pm 0.0024$ \\
\quad w/o FiLM conditioning & $0.4773 \pm 0.0209$ & $0.6557 \pm 0.0130$ & $0.3438 \pm 0.0200$ & $0.3882 \pm 0.0132$ & $0.1174 \pm 0.0018$ & $0.1339 \pm 0.0024$ \\
\quad w/o visual description & $0.4863 \pm 0.0223$ & $0.6571 \pm 0.0153$ & $0.3565 \pm 0.0193$ & $0.4120 \pm 0.0218$ & $0.1206 \pm 0.0031$ & $0.1395 \pm 0.0060$ \\
\quad w/o image & $0.4863 \pm 0.0890$ & $0.6508 \pm 0.0703$ & $0.3255 \pm 0.0924$ & $0.3635 \pm 0.0832$ & $0.1174 \pm 0.0018$ & $0.1339 \pm 0.0024$ \\
\quad text only & $0.4957 \pm 0.0210$ & $0.6678 \pm 0.0134$ & $0.3405 \pm 0.0250$ & $0.3848 \pm 0.0143$ & $0.1174 \pm 0.0018$ & $0.1339 \pm 0.0024$ \\
\quad w/o SupCon loss & $0.4876 \pm 0.0094$ & $0.6618 \pm 0.0058$ & $0.3539 \pm 0.0143$ & $0.3917 \pm 0.0099$ & $0.1174 \pm 0.0018$ & $0.1339 \pm 0.0024$ \\
\quad w/o auxiliary head & $0.4827 \pm 0.0093$ & $0.6566 \pm 0.0093$ & $0.3567 \pm 0.0067$ & $0.3967 \pm 0.0058$ & $0.1174 \pm 0.0018$ & $0.1339 \pm 0.0024$ \\
Single-task (2.1 only) & $0.4850 \pm 0.0172$ & $0.6581 \pm 0.0122$ & -- & -- & -- & -- \\
Single-task (2.2 only) & -- & -- & $0.3648 \pm 0.0214$ & $0.4035 \pm 0.0151$ & -- & -- \\
Single-task (2.3 only) & -- & -- & -- & -- & $0.1346 \pm 0.0246$ & $0.1605 \pm 0.0375$ \\
\hline
\end{tabular}}
\end{table}

\begin{table}[htbp]
\caption{Component ablation on the validation set, soft evaluation on both English and Spanish split. Each cell shows mean $\pm$ standard deviation over five random seeds. All configurations are trained on the non-augmented dataset, using only the deep learning model. Cross-Entropy is not reported for Task 2.3.}
\label{tab:comp_ablation_soft}
\centering
\resizebox{\columnwidth}{!}{%
\begin{tabular}{lcccccc}
\hline
\multirow{2}{*}{\textbf{Configuration}} & \multicolumn{2}{c}{\textbf{Task 2.1}} & \multicolumn{2}{c}{\textbf{Task 2.2}} & \multicolumn{2}{c}{\textbf{Task 2.3}} \\
 & ICM-Soft Norm & CE & ICM-Soft Norm & CE & ICM-Soft Norm & CE \\
\hline
Full system (baseline) & $0.3935 \pm 0.0095$ & $0.9636 \pm 0.0024$ & $0.3342 \pm 0.0078$ & $1.4779 \pm 0.0098$ & $0.1464 \pm 0.0039$ & -- \\
\quad w/o FiLM conditioning & $0.3893 \pm 0.0077$ & $0.9641 \pm 0.0043$ & $0.3306 \pm 0.0067$ & $1.4791 \pm 0.0112$ & $0.1450 \pm 0.0034$ & -- \\
\quad w/o visual description & $0.4031 \pm 0.0084$ & $0.9589 \pm 0.0079$ & $0.3415 \pm 0.0100$ & $1.4710 \pm 0.0116$ & $0.1635 \pm 0.0060$ & -- \\
\quad w/o image & $0.3884 \pm 0.0450$ & $0.9477 \pm 0.0256$ & $0.3319 \pm 0.0273$ & $1.4610 \pm 0.0279$ & $0.1542 \pm 0.0174$ & -- \\
\quad text only & $0.3837 \pm 0.0132$ & $0.9495 \pm 0.0062$ & $0.3278 \pm 0.0095$ & $1.4633 \pm 0.0092$ & $0.1534 \pm 0.0090$ & -- \\
\quad w/o SupCon loss & $0.3944 \pm 0.0099$ & $0.9628 \pm 0.0025$ & $0.3347 \pm 0.0077$ & $1.4773 \pm 0.0099$ & $0.1465 \pm 0.0038$ & -- \\
\quad w/o auxiliary head & $0.3990 \pm 0.0045$ & $0.9610 \pm 0.0045$ & $0.3371 \pm 0.0061$ & $1.4767 \pm 0.0105$ & $0.1481 \pm 0.0034$ & -- \\
Single-task (2.1 only) & $0.3889 \pm 0.0024$ & $0.9606 \pm 0.0060$ & -- & -- & -- & -- \\
Single-task (2.2 only) & -- & -- & $0.3406 \pm 0.0099$ & $1.4701 \pm 0.0122$ & -- & -- \\
Single-task (2.3 only) & -- & -- & -- & -- & $0.1553 \pm 0.0407$ & -- \\
\hline
\end{tabular}}
\end{table}

We applied paired Welch $t$-tests against the baseline configuration for every ablation-metric pair across both hard and soft evaluation, yielding a family of $63$ tests, and correct for multiple comparisons using the Holm-Bonferroni step-down procedure at $\alpha=0.05$. No ablation produced an effect that survives this correction. The closest case is the increase in soft ICM-Soft Norm on Subtask 2.3 when the visual description is removed with uncorrected Welch $p=0.001$, but this fails the Holm threshold ($0.05/63 \approx 0.00079$). We therefore cannot conclude that any individual architectural component carries a statistically detectable benefit at our seed count. The system gains appear to arise from the integration of components rather than any single contribution we can isolate.

Two qualitative observations remain. First, the no-image configuration exhibits dramatically higher variance than every other configuration with standard deviation of $0.089$ on Subtask 2.1 ICM-Norm against $\le 0.023$ for the others. This suggests that the image stream provided training stability even when its mean contribution is small. Second, the multi-task baseline and most ablations of it produce essentially identical performance on Subtask 2.3 with $0.1174 \pm 0.0018$ on hard ICM-Norm. This indicates that the Subtask 2.3 head is collapsing to a near-trivial solution across configurations, only single-task training on 2.3 with ICM-Norm $0.1346 \pm 0.0246$ breaks this pattern, suggesting that multi-task inference, not architectural choices, is the binding constraint on Subtask 2.3 performance.

\subsection{Discussions}
Our results partially support the central hypothesis of this work: annotator physiological and demographic context is statistically associated with the perceived sexism of meme content, and a system that conditions on it stays above the trivial baseline on Subtasks 2.1 and 2.2. At the same time, our multi-seed ablation conditions tempers the claim, when each component is isolated, none produces an effect distinguishable from run-to-run variance after correction. We therefore frame the human-centered conditioning as a useful component of an integrated system rather than as an independently decisive module, and we report the negative ablation openly as a finding.

Comparing the same configuration on validation and test for the All split (i.e., both English and Spanish) with hard ICM-Norm $0.5210$ vs. $0.4861$, the model exhibits modest validation overfitting but no catastrophic distribution shift. The English-Spanish gap on test with $0.5496$ vs $0.4306$ hard ICM-Norm is substantially wider than on validation with $0.5609$ vs $0.4679$, suggesting that the Spanish portion of the test distribution diverges from training more than English portion does, consistent with the multilingual pretraining of our text and vision backbones favoring English idioms of online sexism. The strong relative standing on Subtask 2.2 which is ranked 29th of 114 overall for soft and 71st of 183 for hard evaluation indicates that the cross-attention grounding of the visual description in the image is helpful for reasoning about communicative intent, where both modalities are jointly informative.

The gap between hard and soft performance is also informative. Because we explicitly optimize a soft KL divergence over the annotator label distribution, the model is trained to reproduce disagreement rather than collapse to a majority vote, which is better aligned with the soft evaluation protocol. This is consistent with the learning from disagreement framing \cite{uma2021learning,wu2023don} that motivated our approach.

\subsection{Limitations}
Our approach has several limitations. First, physiological and demographic signals are aggregated by averaging across all annotators of a meme, which discards individual-level variation. A viewer-specific model could better capture the subjectivity the dataset was designed to expose. 

Second, the eye-tracking features exhibit extreme multicollinearity (e.g., Fixations and Saccades with $\rho \approx 1.0$), so the four-dimensional sensor vector likely carries less independent information than its dimensionality suggests. 

Third, EEG signals were not individually significant in our statistical analysis and therefore do not inform the neural branch, entering only through the classical SVM, leaving open the question of whether richer temporal EEG modeling, rather than aggregated bandpower features, might unlock signal that our current setup leaves unused. 

Fourth, our multi-seed ablation (Tables~\ref{tab:comp_ablation_hard} and~\ref{tab:comp_ablation_soft}) finds that no individual architectural component produces an effect distinguishable from run-to-run variance at $n=5$ seeds after Holm-Bonferroni corrections. While this rules out large isolated contributions, smaller real effects could exist below our detection threshold and would require either larger seed counts or a less noisy evaluation regime to confirm.

Fifth, the model class coverage is uneven on the multi-class subtasks, on validation the per-class F1 for the JUDGEMENTAL category from Task 2.2 and for every minority category in Task 2.3 is $0.0$, meaning the model never predicts these categories despite their presence in the training set, broken only by single-task training on 2.3 alone, which indicates that multi-task interference rather than architectural design is the binding constraint on fine-grained categorization performance. This is a symptom of severe class imbalance combined with a soft objective that does not explicitly penalize ignoring rare classes, and explains why our overall Subtask 2.3 ICM-Norm remains close to the trivial baseline.  

Sixth, our cross-lingual augmentation strategy lifts Spanish performance but degrades English seen in Table~\ref{tab:ablation}, a more selective approach, such as translation-quality filtering or back-translation consistency checks, might preserve the benefit without the cost. 

Seventh, our handling of tall, storytelling-format memes relies on square padding, which compresses their content. Splitting such memes into sub-images was not explored and could improve both OCR and visual description quality extraction. Finally, the deep model and the SVM are optimized separately and combined using grid search, rather than trained jointly end-to-end.

\section{Conclusions} \label{sec:conclusions}
We presented a human-centered multimodal system for sexism detection in memes that combines textual, visual, demographic, and physiological modalities through a FiLM-conditioned \cite{perez2018film} cross-attention architecture. Our approach treats sexism identification as a distribution learning problem, using a soft-label KL divergence to capture the inherent subjectivity of human annotators. The system is most competitive on Subtask 2.2 (source intention), where it ranks 29th of 114 overall under soft evaluation, and it stays above the organizers baseline on Subtasks 2.1 and 2.2, confirming that the human-centered conditioning contributes signal rather than noise.

The statistical analysis performed on the EXIST 2026 dataset revealed that annotator physiological responses are significantly associated with the perceived sexism of meme content. By incorporating these signals through a pretrained sensor autoencoder with FiLM modulation, the model learns to adjust its predictions based on the measured cognitive friction that sexist content produces in human viewers. At the same time, our multi-seed ablation under multiple comparison control found that no individual component can be shown to carry a statistically detectable benefit at our seed count. The system signal appears to arise from the integration of the components rather than from any single module. We view this as a useful, honestly reported result for future work that adds physiological and demographic signals to subjective NLP tasks.

In future work, we would like to explore individual-level sensor modeling rather than meme-level aggregation, attention between the Gemma extracted text and image modalities, larger seed counts and lower noise evaluation regimes to resolve the small per-component effects, and adjusting a universal model for all of EXIST Task 2 subtasks that learns from all the labels before being specialized for each task.

\begin{acknowledgments}
The research presented in this paper is supported in part by The Academy of Romanian Scientists, through the funding of the project ``NetGuardAI: Intelligent system for harmful content detection and immunization on social networks'' (AOȘR-TEAMS-IV).
\end{acknowledgments}

%% The declaration on generative AI comes in effect
%% https://ceur-ws.org/GenAI/Policy.html
\section*{Declaration on Generative AI}
  % {\em Either:}\newline
  During the preparation of this work, the authors used Claude Opus 4.6 for rephrasing in order to improve clarity and style. After using this tool, the authors reviewed and edited the content as needed and take full responsibility for the publication’s content.
  % \newline
  
 % \noindent{\em Or (by using the activity taxonomy in ceur-ws.org/genai-tax.html):\newline}
 % During the preparation of this work, the author(s) used X-GPT-4 and Gramby in order to: Grammar and spelling check. Further, the author(s) used X-AI-IMG for figures 3 and 4 in order to: Generate images. After using these tool(s)/service(s), the author(s) reviewed and edited the content as needed and take(s) full responsibility for the publication’s content. 

%%
%% Define the bibliography file to be used
\bibliography{exist.bib}

\appendix

\end{document}